\documentclass[3p]{elsarticle}

\usepackage{lineno,hyperref}
\modulolinenumbers[5]

\usepackage{graphicx}
\usepackage{float}
\usepackage{amsmath}
\usepackage{amssymb}
\usepackage{multirow}
\usepackage{booktabs}
\usepackage{url}
\usepackage{array}
\usepackage{enumitem}
\usepackage{algorithm}
\usepackage{algorithmic}
\usepackage{pifont}
\usepackage[normalem]{ulem}
\usepackage{makecell}
\usepackage{stackengine}
\usepackage{bm}
\usepackage{subfigure}
\usepackage{natbib}

\journal{iScience}

\biboptions{authoryear}

\begin{document}

\begin{frontmatter}

\title{Multilingual Multi-Aspect Explainability Analyses on Machine Reading Comprehension Models}


\author[mymainaddress,mysecondaryaddress]{Yiming Cui\corref{myleadcontact}}
\ead{ymcui@ir.hit.edu.cn}

\author[mymainaddress]{Wei-Nan Zhang}
\author[mymainaddress]{Wanxiang Che}
\author[mymainaddress]{Ting Liu\corref{mycorrespondingauthor}}
\ead{tliu@ir.hit.edu.cn}

\author[mysecondaryaddress]{Zhigang Chen}
\author[mysecondaryaddress,mythirdaddress]{Shijin Wang}

\cortext[mycorrespondingauthor]{Corresponding author}
\cortext[myleadcontact]{Lead contact}

\address[mymainaddress]{Research Center for Social Computing and Information Retrieval, Harbin Institute of Technology, Harbin 150001, China}
\address[mysecondaryaddress]{State Key Laboratory of Cognitive Intelligence, iFLYTEK Research, Beijing 100083, China}
\address[mythirdaddress]{iFLYTEK AI Research (Central China), Wuhan 430000, China}

\begin{abstract}
Achieving human-level performance on some of the Machine Reading Comprehension (MRC) datasets is no longer challenging with the help of powerful Pre-trained Language Models (PLMs). 
However, the internal mechanism of these artifacts remains unclear, placing an obstacle for further understanding these models. 
This paper focuses on conducting a series of analytical experiments to examine the relations between the multi-head self-attention and the final MRC system performance, revealing the potential explainability in PLM-based MRC models. 
To ensure the robustness of the analyses, we perform our experiments in a multilingual way on top of various PLMs. 
We discover that passage-to-question and passage understanding attentions are the most important ones in the question answering process, showing strong correlations to the final performance than other parts.
Through comprehensive visualizations and case studies, we also observe several general findings on the attention maps, which can be helpful to understand how these models solve the questions.
\end{abstract}


\end{frontmatter}


\section*{Introduction}
Teaching machines to read and comprehend human language is an important topic in artificial intelligence (AI).
Machine Reading Comprehension (MRC) has been regarded as an important task to test how well the machine comprehends human languages.
Machine reading comprehension task is to read and comprehend given passages and answer relevant questions, which is a type of Question Answering (QA) task but focuses more on text comprehension.
In the earlier stage, as most of the MRC models \citep{dhingra-etal-2017,kadlec-etal-2016,cui-acl2017-aoa} are solely trained on the training data of individual MRC datasets without much prior knowledge, their performances are not very impressive and are far from humans.
In recent years, the pre-trained language model (PLM) has become a new way for text representation.
Pre-trained language models utilize large-scale text corpora and self-supervised approaches to learn the text semantics.
Various PLMs have bring significant improvements on many natural language processing (NLP) tasks, including BERT \citep{devlin-etal-2019-bert}, RoBERTa \citep{liu2019roberta}, ELECTRA \citep{clark2020electra}, ALBERT \citep{Lan2019ALBERT}, MacBERT \citep{chinese-bert-wwm}, etc.
With the development of PLMs, many MRC models could outperform human performance on a series of MRC benchmarks, such as SQuAD 1.1 \citep{rajpurkar-etal-2016} and SQuAD 2.0 \citep{rajpurkar-etal-2018-know}, indicating that these models can comprehend human languages to a certain extent.

However, achieving human-level prediction performance is not the only goal in AI research.
The decision process and the explanation of these AI models remain unclear, raising concerns about their reliability and placing obstacles to achieving controllable and reliable AI.
In this context, explainable artificial intelligence (XAI) \citep{gunning2017explainable} becomes more important than ever not only in the NLP field but also in various directions of AI.
The goal of XAI is to produce more explainable machine learning (ML) models while preserving a high accuracy of the model prediction.
XAI provides a way for humans to understand the intrinsic mechanism of AI models.
To improve the AI system's explainability, one could seek decomposability of the conventional machine learning model, such as decision trees, rule-based systems, etc.
Moreover, we can also use post-hoc techniques for deep learning models \citep{xai-survey,james-etal-2019}.
However, most of the cutting-edge systems are developed on artificial neural networks, and investigating the explainability of these models is non-trivial.
In this context, some researchers advocate using interpretable models instead of explaining black-box machine learning models \citep{rudin-etal-2019}.
Nonetheless, the community has made great efforts on explaining the neural network model's behavior by post-hoc approaches \citep{cui-etal-2022-teaching}, probing tasks \citep{vulic-etal-2020-probing}, and visualizations \citep{jain-wallace-2019-attention}, etc.

However, understanding the intrinsic mechanism of the neural network is still a challenging issue.
In the NLP field, most of the models rely on the attention mechanism \citep{bahdanau-etal-2014} to model the importance of the input text. 
Later, transformer-based PLMs are becoming a new paradigm to process NLP tasks, whose core component is the multi-head self-attention mechanism \citep{vaswani2017attention}.
While PLMs achieve excellent performance across various NLP tasks, it is necessary to know what is going on inside the multi-head self-attention mechanism.

As a representative PLM, Bidirectional Encoders from Transformers (BERT) \citep{devlin-etal-2019-bert} has become a popular testbed for explainability studies.
Some researchers conducted analyses to help us better understand the internal mechanism of BERT-based architecture.
For example, \citet{kovaleva-etal-2019-revealing} discovered that there are repetitive attention patterns across different heads in the multi-head self-attention mechanism indicating its over-parametrization in BERT. 
Among various research topics on explainability in NLP, perhaps the most trending one is {\em whether the attention can be treated as explanations}. 
Unlike the attention in computer vision area, such as using attention heatmap to visualize how machine understands chest radiograph \citep{Preechakul2022}, the explainability of the attention mechanism is still uncertain in NLP.
Some researchers argue that attention could not be used as explanation.
For example, \citet{jain-wallace-2019-attention} verify that using completely different attention weights could also achieve the same prediction.
However, on the contrary, some works hold positive attitudes about this topic, and they believe that the attention mechanism is a source of explainability \citep{wiegreffe-pinter-2019-attention,bastings-filippova-2020-elephant}.
These works have brought us various views on the attention mechanism in PLMs, but there is still no consensus about this important topic as of now.
Also, most of these works only investigate the text classification tasks, which require less reasoning skills and lack a comprehensive understanding of the long text.

Regarding the explainability studies in MRC tasks, \citet{yang-etal-2018-hotpotqa} proposed a multi-hop question answering dataset, called HotpotQA.
However, unfortunately, most of its following works only focus on improving the system performance without specifically caring about the explainability.
\citet{cui-etal-2022-teaching} proposed an unsupervised approach to extract evidence span in the passage, which can be seen as a post-hoc explanation.
\citet{cui-etal-2021-expmrc} proposed a comprehensive benchmark for evaluating the explanations in MRC tasks, including span-extraction MRC and multi-choice MRC for both English and Chinese.
However, most of these works mainly focus on the post-hoc explainability approaches, which lack a comprehensive understanding of the internal mechanism of the model itself.
\citet{wu-etal-2021-evaluating} investigated several black-box attacks at the character, word, and sentence level for MRC systems.
Overall, a comprehensive and robust explainability investigation on the MRC model is not well-studied in the previous literature.

To increase the diversity in better understanding the attention mechanism in PLMs, in this paper, we present an explanatory study specifically for the MRC tasks. 
Except for the traditional attention visualizations in a layer-wise or head-wise view, we also provide a thorough view with extensive and robust experiments to better understand whether these observations can be generalizable to other PLMs and even for the PLMs in a different language or size. 
Our contributions are listed as follows.
\begin{itemize}
	\item We specifically aim to investigate the attention mechanism of PLM-based MRC models in various aspects of the PLMs, including the language, model type, capacity, etc. As far as we know, this is the first work that analyzes the MRC model's explainability in a multilingual and multi-aspect way.
	\item Through massive analytical experiments, we find that {\em passage-to-question} and {\em passage understanding} attention are the most important zones in the attention map, which might be the sources for the model's explainability.
	\item Several interesting observations are discovered, including model-specific behaviors in attention map, etc., which can be useful in better understanding the internal mechanism of these MRC models when solving the questions.
\end{itemize}

\section*{Results}\label{sec-experiments}

\subsection*{A New View on Attention Map: Attention Zones}
Before presenting our analyses, we first present a new view on the attention map in MRC models, which is a crucial component throughout this paper.
Formally, MRC tasks consist of three essential parts: passage $\bm{P}$, question $\bm{Q}$, and answer $\bm{A}$.
Usually, we concatenate the passage and the question into the pre-trained language model, letting them interact with each other, and finally, the model outputs an answer. 
Specifically, the input is organized as follows, where the {\tt [CLS]} represents the special starting token, {\tt [SEP]} represents the special separating token, respectively.
\begin{equation}\nonumber
	\texttt{[CLS]}~~\text{Question}~~\texttt{[SEP]}~~\text{Passage}~~\texttt{[SEP]} 
\end{equation}

\begin{figure*}[htbp]
  \centering
  \includegraphics[width=1.0\textwidth]{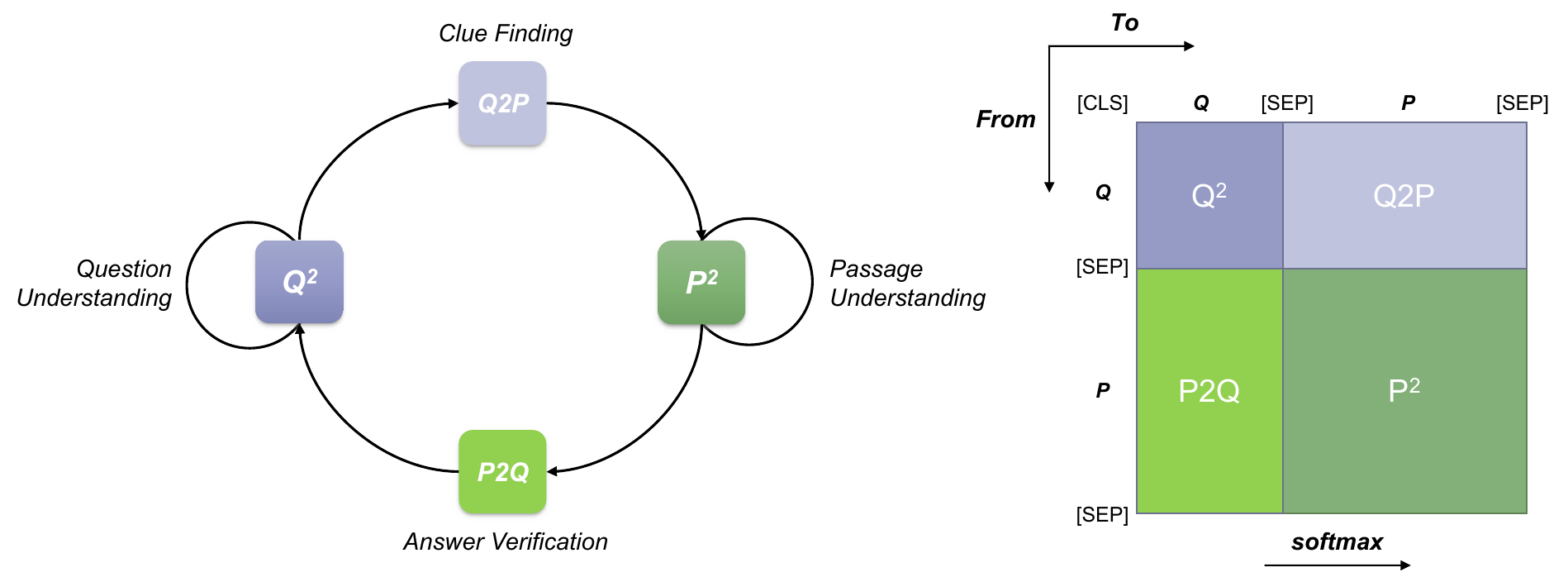}
  \caption{\label{attention-zones} Intuitive explanation of four different attention zones for MRC tasks. Q: Question, P: Passage.} 
\end{figure*}

Unlike previous works that regard the attention map as a whole, in this paper, we propose to decompose the attention map in a much more precise view, which is specifically designed for MRC tasks.
To have a better understanding of the multi-head self-attention in MRC models, we divide the attention map $\bm{M} \in\mathbb{R}^{L \times L}$ into four areas (where $L$ is the length of the input), namely {\em attention zones}, as shown in Figure \ref{attention-zones}.
For each part, we give intuitive illustrations as follows.\footnote{These illustrations may not represent the actual behavior in transformer model but can help us understand them intuitively.}
\begin{itemize}[itemsep=2pt,topsep=2pt,parsep=4pt]
	\item {\bf Q$^2$}: The question is attended to itself, which can be seen as {\em question understanding} process.
	\item {\bf Q2P}: It represents the distributions of passage words in terms of a specific question word, which can be seen as {\em finding clues using the question} process.
	\item {\bf P2Q}: Similar to the Q2P, but in a reverse order, which can be seen as {\em answer verification} process.
	\item {\bf P$^2$}: The passage text is attended to itself, which can be seen as {\em passage understanding} process.
\end{itemize}

In the following sections, we observe the behaviors in different attention zones rather than regard the attention map as a whole.
This allows us to understand the attention mechanism in MRC models better.

\subsection*{Experimental Setups}\label{experimental-setups}
In this paper, we aim to analyze the span-extraction MRC, which is one of the most representative MRC tasks.
The span-extraction MRC task is to read a passage and answer the relevant question, where the answer is an exact span in the passage.
Specifically, we use SQuAD \citep{rajpurkar-etal-2016} dataset for English and CMRC 2018 \citep{cui-emnlp2019-cmrc2018} dataset for Chinese to simultaneously evaluate attention behaviors in both languages.
To build MRC models, following previous works, we use BERT \citep{devlin-etal-2019-bert} as a natural baseline for most of the experiments.
We use BERT-base-cased model\footnote{\url{https://storage.googleapis.com/bert_models/2018_10_18/cased_L-12_H-768_A-12.zip}} for English and BERT-base\footnote{\url{https://storage.googleapis.com/bert_models/2018_11_03/chinese_L-12_H-768_A-12.zip}} for Chinese for weight initialization.

Unlike most previous works that only report single-run experimental results, to make our observations more robust and reliable, all experiments are trained and evaluated five times (with different random seeds), and their average scores are reported.

\subsection*{Quantitative Study: Attention is Conditional Explanation}
\begin{figure}[h]
  \centering
  \subfigure{\includegraphics[width=0.95\columnwidth]{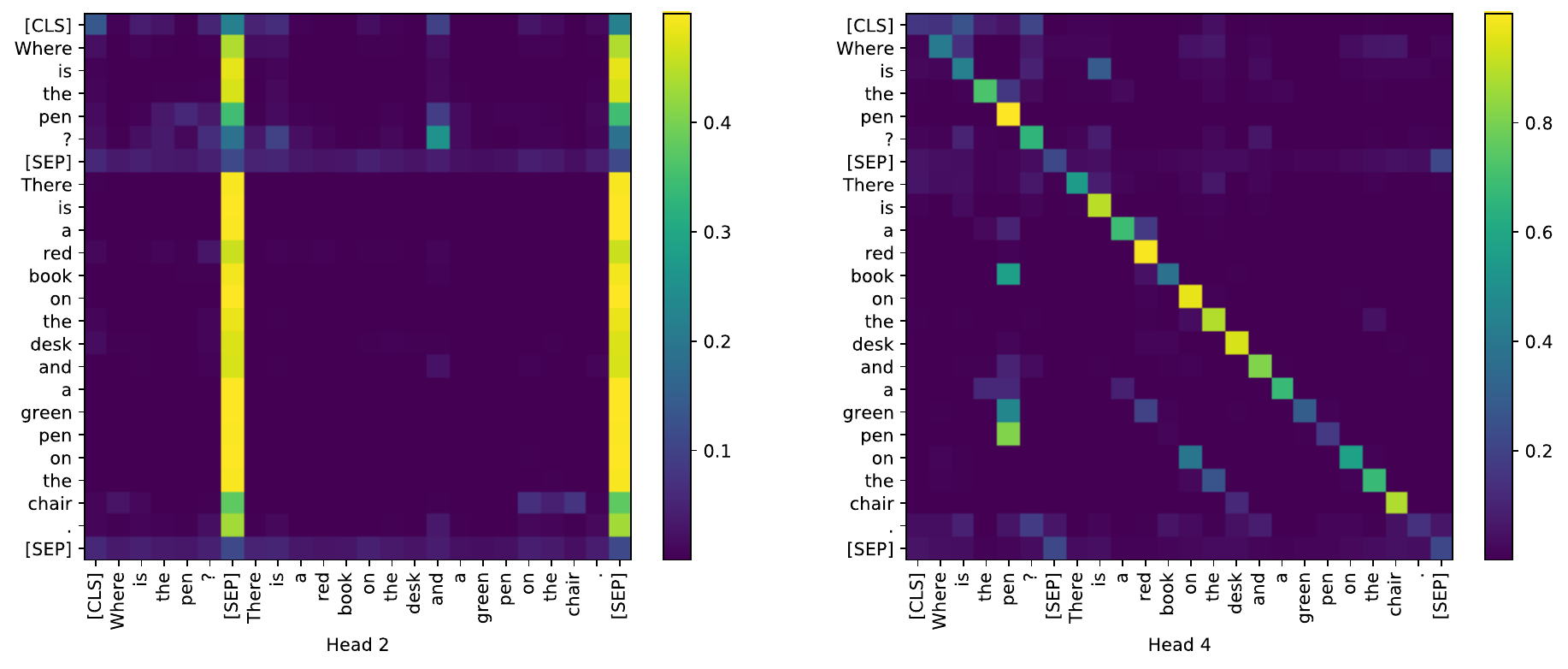}}
  \caption{\label{special-diagonal} Attention maps of 2nd and 4th head in the last layer of fine-tuned BERT$_\text{base}$ on SQuAD. There are strong patterns in diagonal elements and the elements that related to special tokens.} 
\end{figure}

Firstly, we investigate the effect of masking some parts in the attention map.
Recall that the PLMs show a strong pattern for special tokens ({\tt [CLS]} and {\tt [SEP]}) and diagonal tokens in the attention map \citep{kovaleva-etal-2019-revealing}, as shown in Figure \ref{special-diagonal}.
To examine the effect of these tokens, we mask the special tokens or attention zones during the training phase to see their dependence on the model performance.
Masking means to set the whole (or partial) attention map with all the same and large negative numbers (say -10000). 
After the softmax function, the resulting attention map loses its ability to ``highlight'' important relations in the respective area.
The results are shown in Table \ref{result-masking}.

\begin{table}[h]
\begin{center}
\caption{\label{result-masking} Results on masking part of attention map. } 
\begin{tabular}{l c c c c}
\toprule
& \multicolumn{2}{c}{\bf SQuAD} & \multicolumn{2}{c}{\bf CMRC 2018} \\
& \bf EM & \bf F1 & \bf EM & \bf F1 \\
\midrule
Baseline		& 80.687 & 88.129 & 63.796 & 84.789 \\
\midrule
No \tt[CLS]  	& 80.802 & 88.276 & 64.119 & 84.858 \\
No Mid \tt[SEP] 	& 80.689 & 88.082 & 63.896 & 84.626 \\
No End \tt[SEP]	& 80.522 & 87.959 & 64.299 & 84.866 \\
No All			& 78.956 & 86.414 & 63.659 & 83.945 \\
\midrule
No Diagonal		& 80.645 & 88.241 & 64.548 & 84.908 \\ 
\midrule
No Q$^2$		& 76.395 & 84.195 & 60.100 & 80.625 \\
No Q2P			& 79.941 & 87.352 & 64.517 & 84.592 \\
No P2Q			& 12.763 & 16.355 & 15.070 & 18.466 \\	
No P$^2$		& 34.441 & 51.792 & 16.278 & 42.906 \\
\bottomrule
\end{tabular}
\end{center}
\end{table}

We first look into the results of masking special tokens.
Not surprisingly, removing all three special tokens yields a decline in the performance, where it hurts more on SQuAD than CMRC 2018.
However, when removing these tokens individually, we did not see a significant drop and even noticed a rise in the performance, such as removing {\tt [CLS]} in CMRC 2018.
Also, when removing diagonal elements, we see a relatively consistent improvement on both datasets, where there is +0.752 EM on CMRC 2018.
This indicates that removing higher attention values does not necessarily result in a significant performance drop.
When masking specific attention zones, we can see that their performances vary a lot.
Removing P$^2$ and P2Q zones hurts the performance most, while there is no significant drop for the Q2P and Q$^2$ zones. 
This demonstrates that the ``{\em from passage to X}'' attentions are relatively more important than ``{\em from question to X}'' in MRC models.
Further discussion will be presented in the next section. 

Secondly, we also present the baseline performance when removing the whole or all top-10 elements in each attention zones for all layers.
This experiment examines whether there is a significant performance drop for a regular baseline system when a certain attention zone is disabled.
As we can see from Table \ref{evidence-analysis}, removing the P2Q zone (partial or whole) hurts performance the most, indicating that the key to answer the questions mostly resides in this attention zone.
On the contrary, the Q2P zone hurts performance least. 
This is in line with the observations in Table \ref{result-masking}.
\begin{table}[h]
\begin{center}
\caption{\label{evidence-analysis} Results on removing Top-$k$ and all attention scores. Only EM scores are reported.} 
\begin{tabular}{l c c c c}
\toprule
& \multicolumn{2}{c}{\bf SQuAD} & \multicolumn{2}{c}{\bf CMRC 2018} \\
& \bf All & \bf Top-10 & \bf All & \bf Top-10  \\
\midrule
All		 & - & 66.539 & - & 57.813 \\
Q$^2$	 & 40.464 & 65.272 & 27.145 & 58.652 \\
Q2P 	 & 77.533 & 79.743 & 56.390 & 63.324 \\
P2Q 	 & 4.354  & 45.790 & 1.634  & 43.939 \\
P$^2$  	 & 6.923  & 78.412 & 28.565 & 63.175 \\
\bottomrule
\end{tabular}
\end{center}
\end{table}

Lastly, as most of the previous works use attention scores to present where the model emphasizes, we wonder whether there is a high correlation between the attention score and system performance.
In this experiment, we mask top-$k^\text{th}$ value in different attention zone and calculate the Pearson correlation between its performance and the rank $k \in \{1\dots10\}$.
\begin{table}[h]
\begin{center}
\caption{\label{results-correlation} Pearson correlation of masking top-$k^\text{th}$ attention score. We report five-run average and its standard deviations.} 
\begin{tabular}{p{1cm} p{2.2cm}<{\centering} p{2.2cm}<{\centering} }
\toprule
& \bf SQuAD & \bf CMRC 2018 \\
\midrule
Q$^2$	& 0.624 $\pm$ 0.083 & -0.316 $\pm$ 0.370 \\
Q2P & 0.159 $\pm$ 0.435 & 0.134 $\pm$ 0.531 \\
P2Q & 0.765 $\pm$ 0.017 & 0.778 $\pm$ 0.118  \\
P$^2$  & 0.534 $\pm$ 0.216 & 0.291 $\pm$ 0.299 \\
\bottomrule
\end{tabular}
\end{center}
\end{table}

As we can see from Table \ref{results-correlation}, not all attention zones correlate well to the system performance.
We can see a consistent higher correlation in P2Q and P$^2$ zones while lower in Q2P zones.
This strengthens our claim that a higher attention score does not necessarily contribute more to the performance.
This also indicates that rather than perform a rough analysis on the whole attention map, it is necessary to conduct experiments on different attention zones, especially those with higher correlations.
Through the experiments above, we conclude that the {\em attention is conditional explanation} in MRC models.
Based on these observations, we proceed with further and deeper analyses on different attention zones to examine their behaviors individually in the rest of the paper.

\subsection*{P2Q and P$^2$ Zones Matter Most in MRC Models}\label{zone-comparison}
Based on the observations in the previous section,  we analyze the attention behavior in different zones in terms of different aspects.
We use the baseline system and experimental setups in the previous section and decode them under different settings.
To make the visualization results comparable, we get the decoding performance (only EM scores are considered) when masking a certain attention zone and calculate the difference to the baseline score for all experiments.
Then we observe the attention behavior in different layers and attention heads in terms of different languages, model's capacity, etc.

\begin{figure}[h]
  \centering
  \includegraphics[width=0.6\columnwidth]{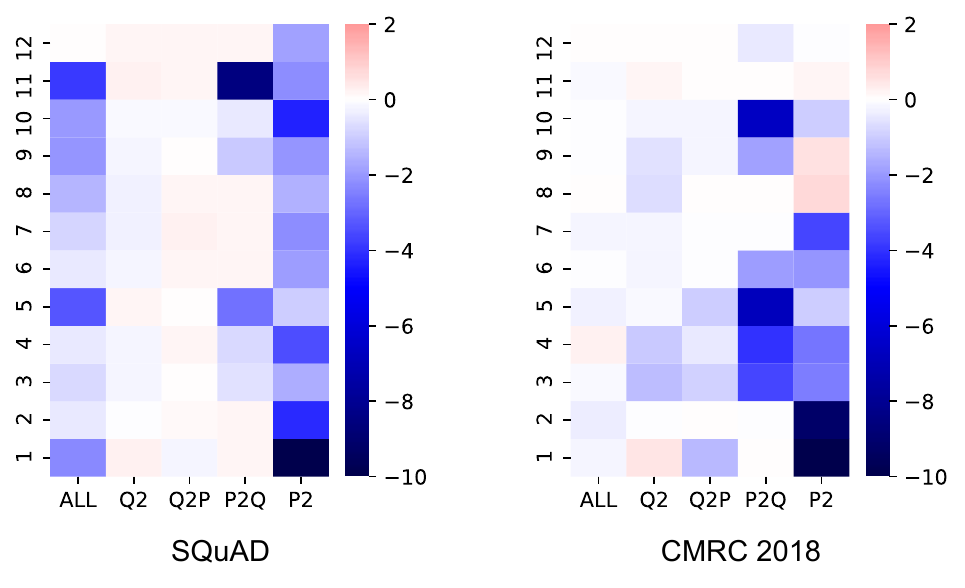}
  \caption{\label{analysis-layerwise} Layer-wise analyses in different attention zones for SQuAD and CMRC 2018. The lighter color means the performance is near the baseline, while darker color means a bigger gap to the baseline (red: above baseline, blue: below baseline).} 
\end{figure}

Firstly, we look into general situations that disable a certain attention zone in a specific attention head or layer.
The layer-wise analysis is depicted in Figure \ref{analysis-layerwise}.
Surprisingly, we find that though the visualizations are made with different datasets and pre-trained language models, two figures look similar in their performance distributions, where we conclude as follows.
\begin{itemize}[itemsep=2pt,topsep=2pt,parsep=4pt]
	\item Disabling Q$^2$ and Q2P does not show significant drops to the overall performance.  
	\item Passage understanding (P$^2$) starts from the first layer and shows a strong reaction after masking. 
	\item Removing the top-most layer (layer-12 in this case) does not show significant performance drops. This is in accordance with the findings that using the representation of the second most layer for fine-tuning results in a better performance on downstream tasks \citep{xiao2018bertservice}. 
\end{itemize}

\begin{figure}[h]
  \centering
  \includegraphics[width=0.8\columnwidth]{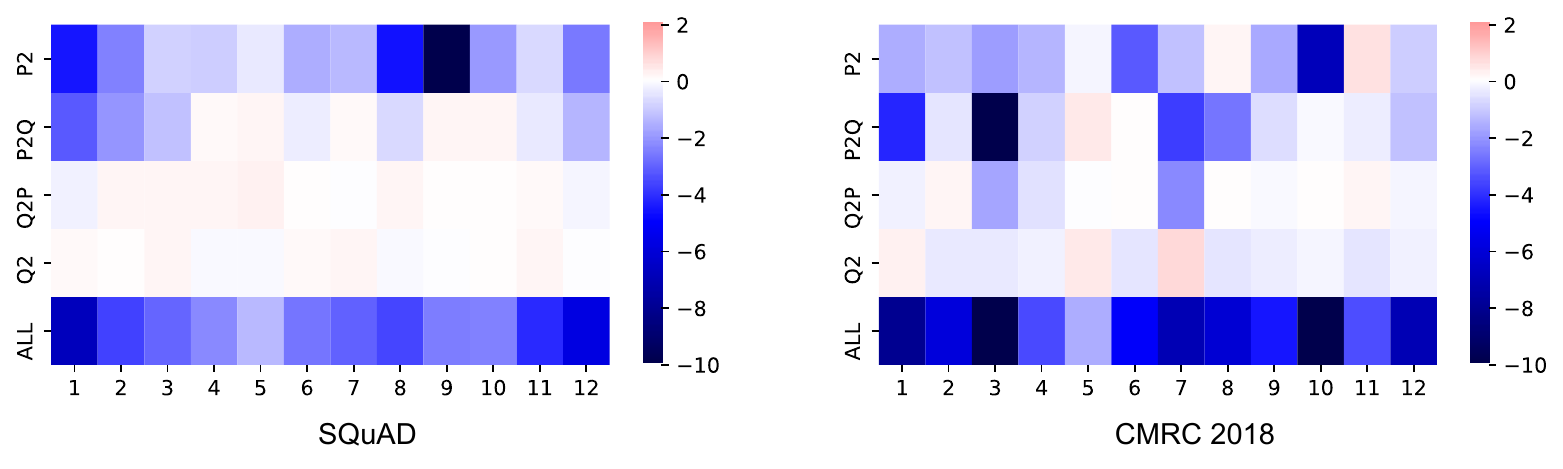}
  \caption{\label{analysis-headwise} Head-wise analyses in different attention zones for SQuAD and CMRC 2018. } 
\end{figure}

We move onto the the head-wise analysis, which is depicted in Figure \ref{analysis-headwise}.
Except for the strong dependence on P$^2$ and P2Q zones, the head-wise view does not show a consistent pattern in SQuAD and CMRC 2018, and thus we focus on layer-wise analyses in the following parts.

Overall, based on the visualizations of layer-wise and head-wise analyses, we induce that the model first pays more attention to modeling the passage itself (P$^2$) to fully understand the text.
Moreover, the question information also flows to the passage (Q2P) to indicate where to attend. 

Next, a natural question would be {\em why removing P2Q attention results in a severe performance drop than Q2P?}
Both P2Q and Q2P are in the shape of $L_{p} \times L_{q}$, and thus it is nothing to do with the area. 
However, if we take a closer look into their positions in the attention map (Figure \ref{attention-zones}), we might possibly understand its reason intuitively. 
The most important thing to keep in mind is that the softmax function is applied in a row-wise manner.
Disabling Q2P indicates that ``{\em from question to passage}'' attention is removed. 
The question can only be attended to itself, and the passage can be attended to both passage itself and the question.
On the contrary, disabling P2Q indicates that the question can be attended to both question itself and passage, but the passage can only be attended to itself.
In MRC tasks, the length of the question (dozens of words) are typically shorter than the length of the passage (several hundreds of words or more). 
In this context, discarding the P2Q zone harms the performance a lot due to the fact that a large amount of the softmax functions cannot be applied to both the question and passage, resulting in an insufficient interaction between them, which is a crucial process in machine reading comprehension.

\subsection*{Harder Questions Requires Deeper Understanding in Question}
\begin{figure*}[h]
\centering
\includegraphics[width=1.0\columnwidth]{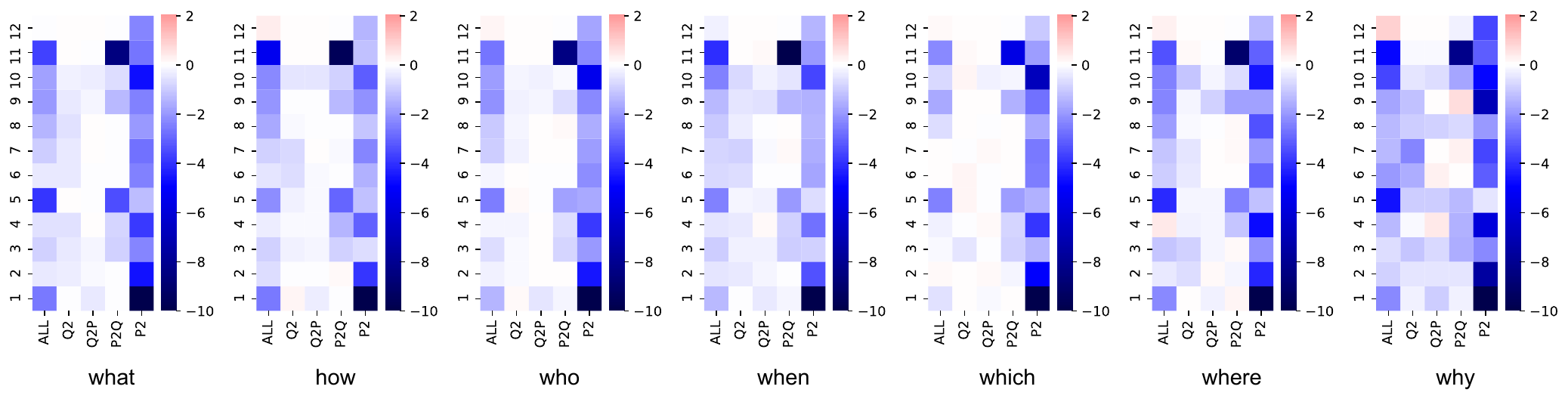}
\caption{\label{analysis-ques-type} Analyses of different question types for SQuAD. The number of each question type (in order): what (6073), how (1389), who (1377), when (864), which (747), where (508), why (158). } 
\end{figure*}

We further analyze the attention behavior for different types of questions, which can help us understand their behaviors in a linguistic view.
We select the seven most frequent question types in SQuAD: {\em what, how, who, when, which, where}, and {\em why}.
The visualizations are shown in Figure \ref{analysis-ques-type}.

As we can see, the attention patterns for different types of questions are quite similar and are similar to the overall attention pattern (Figure \ref{analysis-layerwise}).
Regarding the attention map for ``why'' questions, all zones show stronger impacts on the performance than other types of questions.
Especially, it put more emphasis on the Q$^2$ and P$^2$ attention zones.
As ``why'' questions are relatively harder than the others, the visualization indicates that when solving harder questions, the model focuses more on the question understanding (Q$^2$) and passage understanding (P$^2$), which is in line with problem solving process in human view.

\begin{figure}[h]
  \centering
  \includegraphics[width=0.6\columnwidth]{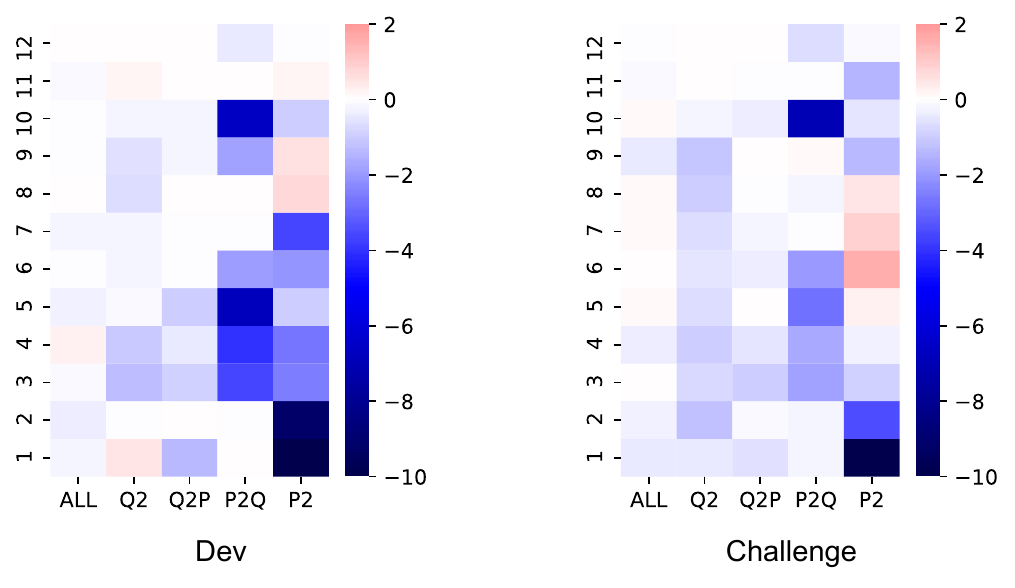}
  \caption{\label{cmrc2018-chl} Comparison of the development and challenge set for CMRC 2018. } 
\end{figure}

Furthermore, CMRC 2018 provides an additional challenge set, which contains the questions that need comprehensive reasoning over multiple sentences.
We can also compare the attention map between the normal development set and the challenge set.
The results are shown in Figure \ref{cmrc2018-chl}.
As we can see, the two figures are quite similar, where the P$^2$ and P2Q are the most important attention zones.
We also discover a stronger focus in the Q$^2$ zone of the challenge set compared to the counterpart.
This observation is similar to Figure \ref{analysis-ques-type} (`{\em why}' questions in SQuAD).
Through the visualizations of both languages, the results strengthen our claims that these hard questions (and longer question text\footnote{The average length of challenge question is 18 compared to 15 in dev set, described in \citet{cui-emnlp2019-cmrc2018}.}) require a deeper understanding of the question in both English and Chinese MRC tasks.

\subsection*{PLM-specific Attention Behaviours}
In the previous analyses, we have observed several interesting and consistent findings. 
However, {\em are these observations generalizable to other PLMs as well}?
To investigate this question, we perform layer-wise decomposition on another two popular PLMs: ELECTRA \citep{clark2020electra} and ALBERT \citep{Lan2019ALBERT}.
Besides, we also carry out experiments on their large-level model ($\sim$340M parameters) to compare with their base-level model ($\sim$110M params).
\begin{itemize} 
	\item {\bf ELECTRA} \citep{clark2020electra} employs a new generator-discriminator framework that is different from most of the previous PLMs. The generator is typically a small masked language model (MLM) that learns to predict the original words of the masked tokens. The discriminator is trained to discriminate whether the input token is replaced by the generator. In the fine-tuning stage, only the discriminator is used.
	\item {\bf ALBERT} \citep{Lan2019ALBERT} mainly focuses on designing a compact PLM by introducing two techniques of parameter reduction. The first is the factorized embedding parameterization, which decomposes the embedding matrix into two small matrices. The second one is the cross-layer parameter sharing in the transformer, which significantly reduces the number of parameters. Besides, they also proposed the sentence order prediction (SOP) task to replace the next sentence prediction (NSP).
\end{itemize}
\begin{figure*}[h]
  \centering
  \includegraphics[width=1.0\columnwidth]{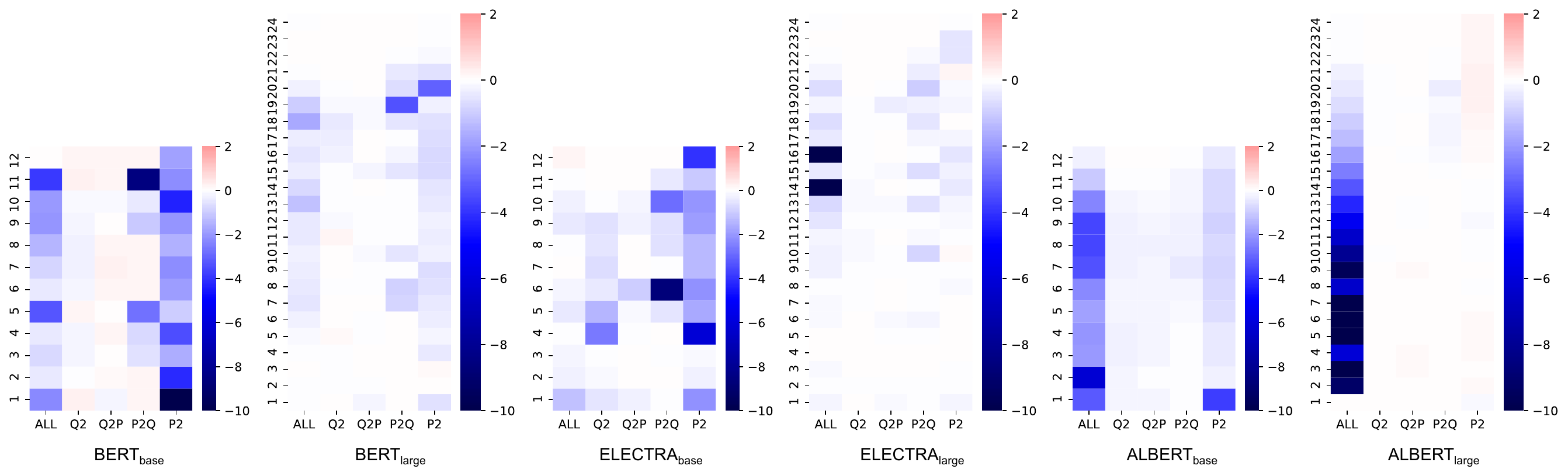}
  \caption{\label{diff-plm} Analyses of base-level and large-level BERT, ELECTRA, ALBERT for SQuAD. } 
\end{figure*}

The results are shown in Figure \ref{diff-plm}.
Overall, the base-level models are much sensitive to the elimination of specific attention zones in several layers.
On the contrary, all large-level models yield minor performance loss (depicted in lighter colors) than the counterpart, which indicates that the large-level models are more robust, and the learning of the model is not concentrated to a few attention zones.
A possible guess is that with a larger capacity for large-level PLMs, there is redundant knowledge stored in the model.
In this way, when a specific attention zone is disabled, the model can still recover such knowledge in other relevant zones, and thus the final performance is not affected that much.
By comparing different PLMs, the importance for different zones are as follows.
\begin{itemize} 
	\item {\bf BERT}: P$^2$ $>$ P2Q $>$ Q$^2$ $\approx$ Q2P
	\item {\bf ELECTRA}: P$^2$ $>$ P2Q $>$ Q$^2$ $>$ Q2P
	\item {\bf ALBERT}: P$^2$ $>$ P2Q $\approx$ Q2P $\approx$ Q$^2$
\end{itemize}

As we can see, P$^2$ and P2Q are the most important attention zones across different sizes and types of PLMs.
However, these PLMs show different attention patterns, indicating their distinct ways of processing the text.
For BERT and ELECTRA, P$^2$ is the most important attention zone, followed by P2Q.
While for ALBERT, it can be seen that the importance of attention is evenly distributed in each attention head and layer.
The main difference between ALBERT and other PLMs is that the parameter of each transformer layer is shared.
Thus, the learning for each attention zones is amortized, as changing the parameter in one layer will also change the attention behavior in other layers. 
In this way, the model could not focus on learning a specific feature at a particular layer or attention head and must be amortized through all layers.
Apart from the observations above, we also notice that 
\begin{itemize}
	\item For base-level PLMs, the passage understanding is mostly learned from the bottom layer, but it still progressively learns in the following layers.
	\item Disabling all attentions in the top layers yields no performance drop and even a minor gain, indicating that there are redundant attention heads that can be pruned without hurting the system performance. 
	\item Disabling all attentions does not necessarily result in worse performance compared to disabling a specific attention zone, and vice versa, such as in the 6th layer of ELECTRA$_\text{base}$ and the 14th and 16th layer of ELECTRA$_\text{large}$. This indicates the interaction complexities between different attention zones.
\end{itemize}

\section*{Discussion}\label{sec-case-study}
In the previous sections, we perform quantitative analyses on the proposed attention zones to explore their behaviors in MRC models.
To further investigate how these attention zones affect the machine reading process, in this section, we come back to visualize the attention map and look into specific examples to analyze the potential behavior of MRC models.
Based on our findings in the previous sections, we visualize the multi-head self-attention to explicitly discover how the model processes the MRC example.
To make the visualization clear, we discard attention values that connect to {\tt [CLS]} and two {\tt [SEP]} special tokens, which have great attention values but do not provide helpful hints on understanding the explainability of the MRC model.

\begin{figure}[h]
  \centering
  \includegraphics[width=1\columnwidth]{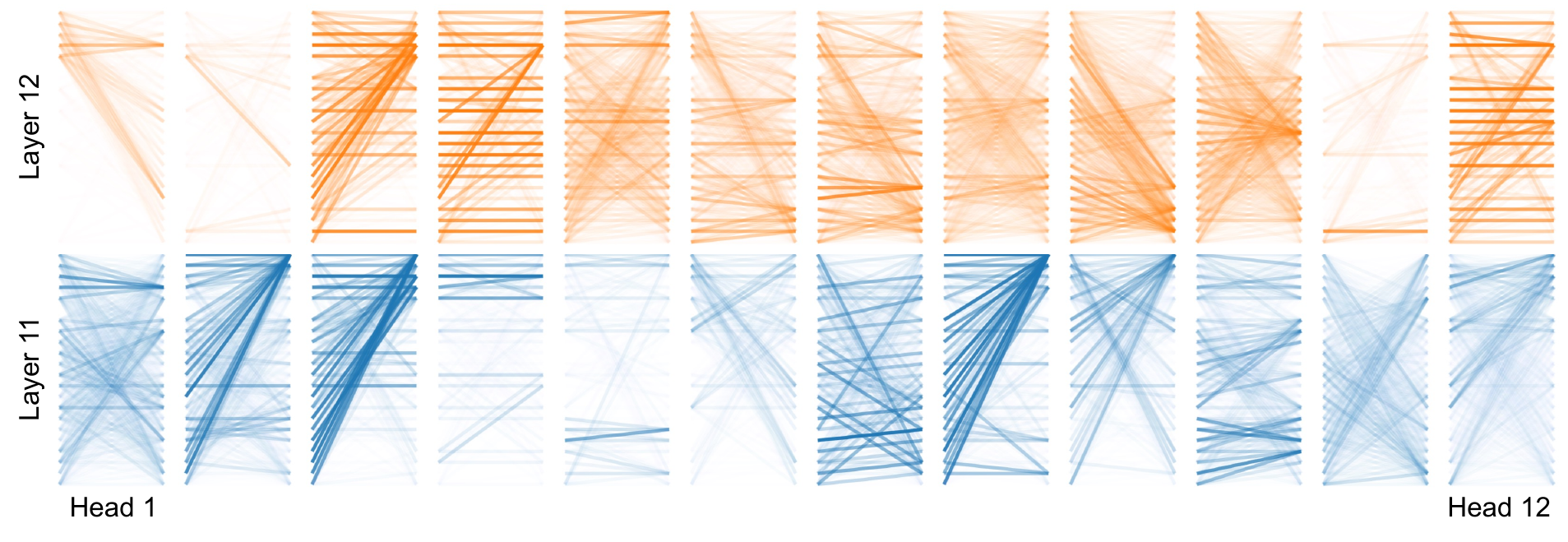}
  \caption{\label{attention-11-12} Attention maps for different attention heads in layer 11 and 12. Recall that the input is created by the concatenation of ``[CLS] Q [SEP] P [SEP]''. The darker line means a strong connection between two words. } 
\end{figure}

Here we use a simple example to examine the attention behavior. 
The passage is ``{\em There is a red book on the desk and a green pen on the chair.}'', and the question is ``{\em Where is the pen?}''.
We omit the full picture of attention maps in all transformer layers and only show the 11th and 12th layer of BERT$_\text{base}$ trained on SQuAD, as shown in Figure \ref{attention-11-12}.
It can be seen that the attention patterns are not fixed for a specific head. 
For example, the 12th head shows a strong ``{\em all-to-question}'' pattern in layer 11, where the majority of the lines are connected to the top right. 
In contrast, it shows an ``{\em identity-mapping}'' in layer 12, where there are many horizontal lines, meaning the words are connected to itself.
Thus, it is not feasible to select a fixed set of attention heads for explainability evaluation across different layers. 

\begin{figure}[h]
\centering
\includegraphics[width=0.7\columnwidth]{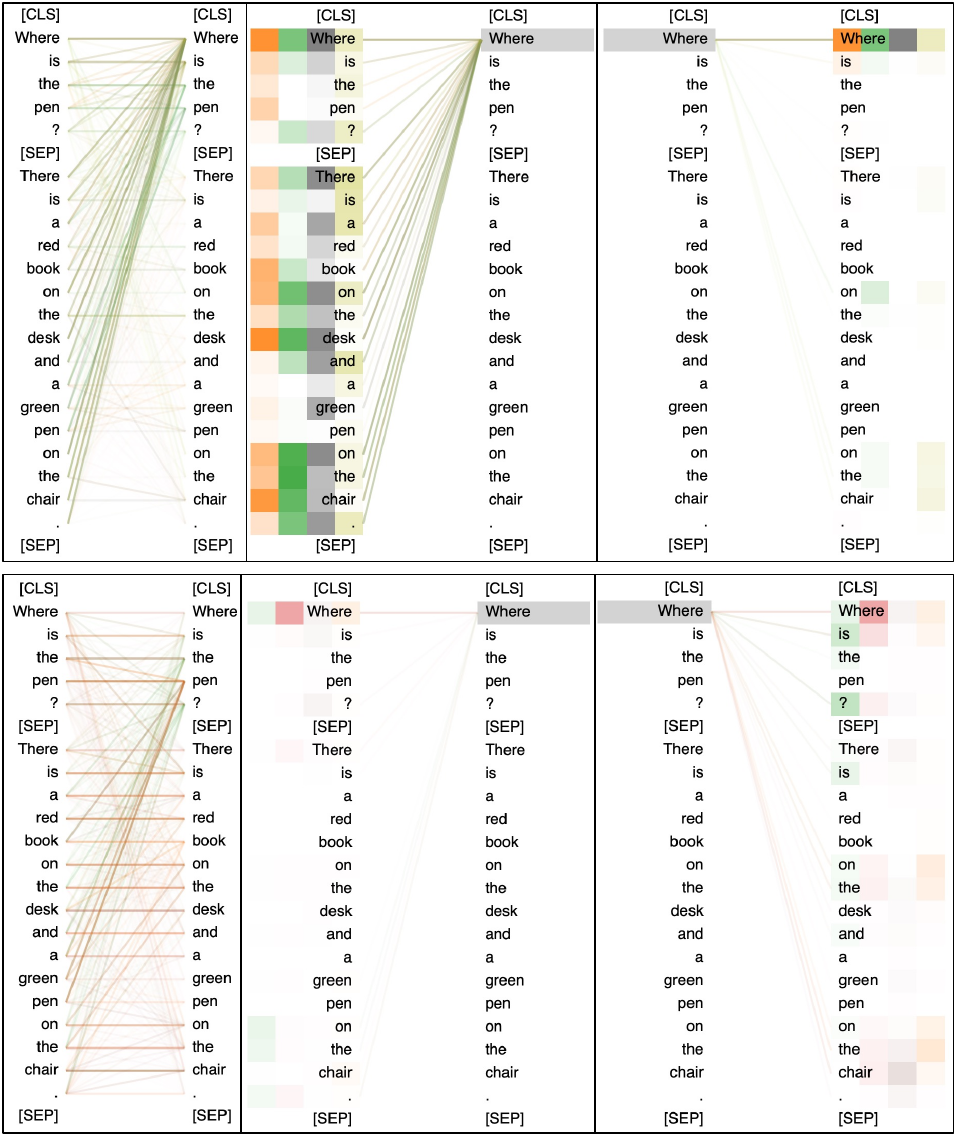}
\caption{\label{attention-11-12-heads} Visualization of specific attention heads in layer 11 (upper) and 12 (lower) of SQuAD model. The four boxes behind each word represents the values of each attention head (the higher value the darker). } 
\end{figure}

In this context, to get a closer look, we manually select the 3-4-8-9th heads for layer 11 and 3-4-6-12th heads for layer 12 to present how the model solves the question in MRC.
We visualize the attention distribution in terms of the question word ``{\em why}'' in attention source and target, which is depicted in Figure \ref{attention-11-12-heads}.
Through observations, we found a strong indication of explainability connected to the question word. 
By comparing the attentions in the 11th and 12th layer, we can see that the question word ``{\em why}'' shows strong attention to other words in layer 11 while it gets weak in layer 12.
However, in layer 12, we can see that the phrase ``{\em on the chair}'' in the passage attends to the question word ``{\em where}'' (P2Q), and the word ``{\em where}'' attends to both ``{\em on the desk}'' and ``{\em on the chair}'' (Q2P). \footnote{Here, we mean a relatively higher attention value than the others.}
We induce that the answer is obtained by taking both P2Q and Q2P attention zones into account, and thus the model champions the phrase ``{\em on the chair}'' as the final answer.
This is also observed in the previous section that removing attentions in layer 12 does not yield significant performance drops, as a similar pattern can also be observed in layer 11.

Besides, another interesting observation is that the attention values are not only higher for the start and end position of the answer span but also the words between them.
It can be seen that the word ``{\em why}'' does not only attend to the word ``{\em on}'' and ``{\em chair}'', but the whole phrase ``{\em on the chair}'' in both layer 11 and 12.
This implies that though the answer span is extracted by the start and end pointers, the MRC models are capable of considering the words between them to make final answer predictions, but not solely on the start and end tokens.
Perhaps this is why almost all span-extraction MRC systems are not modeled in a sequence tagging manner, as the words between start and end positions are already considered in the transformer.

Based on our findings, there might be two directions for future works.
First, we will try to find many-to-many mappings in the attention map, which is much important to the questions that need comprehensive reasoning.
Also, we will find a way to automatically discard the attention head that contributes less to the final system performance, as not all attention heads are important in transformer models.

\section*{Limitations of the Study}\label{sec-limitation}
In this paper, we have discussed the potential explainability within machine reading comprehension models.
Though we have strived to make our analyses as comprehensive as possible, there may have several limitations that need to be studied in future work.
\begin{itemize}
	\item Explainability for other languages: We have studied the explainability in both English and Chinese models, which is a step forward to increase the language diversity, as these two languages belong to different language genres. The conclusions made in this paper might have good generalizations than monolingual experiments. However, it is not sure whether our analytical conclusions are generalizable to other genres of languages, such as Arabic.
	\item Explainability for other models: Though we have found several common phenomena as shown in the visualizations, different pre-trained language models exhibit different patterns in the attention map, especially for those with different neural architecture (such as BERT v.s. ALBERT). It is interesting to see how other PLMs perform in a similar context.
	\item Different ways to examine the attention mechanism: Using attention values or importance scores has been a normal way to visualize the attention map. This paper provides a different way to examine the attention map by using system performance. With the development of XAI studies, it is promising to have a more efficient way to analyze the attention map.
\end{itemize}

As the explainability for machine learning approaches is still an ongoing research topic, we hope that such limitations can be further studied in future work to help us better understand the internal mechanism of machine reading comprehension models.

\section*{Acknowledgments}\label{sec-acknowledgments}
We would like to thank Dr. Kanudha Sharda (Associate Editor) and three anonymous reviewers for their constructive comments to improve our paper. 
This work is supported by the National Key Research and Development Program of China via grant No. 2018YFB1005100, and Natural Science Foundation of Heilongjiang Province of China via grant No. YQ2021F006.
Yiming Cui would like to thank TPU Research Cloud (TRC) program for Cloud TPU access.

\section*{Author Contributions}\label{sec-author-contributions}
Conceptualization, Y.C.;
Methodology, Y.C.;
Formal Analysis, Y.C., W.Z., W.C., S.W.;
Writing - Original Draft, Y.C.;
Writing - Review \& Editing, Y.C., W.Z., W.C., S.W.;
Visualization, Y.C.;
Supervision, T.L.;
Funding Acquisition, Z.C., S.W.;
All authors read and approved the submission of this paper.

\section*{Declaration of Interests}\label{sec-declaration}
The authors declare no competing interests.

\section*{Main Figure Titles and Legends}
Figure 1. Intuitive explanation of four different attention zones for MRC tasks. Q: Question, P: Passage.

Figure 2. Attention maps of 2nd and 4th head in the last layer of fine-tuned BERT$_\text{base}$ on SQuAD. There are strong patterns in diagonal elements and the elements that related to special tokens.

Figure 3. Layer-wise analyses in different attention zones for SQuAD and CMRC 2018. The lighter color means the performance is near the baseline, while darker color means a bigger gap to the baseline (red: above baseline, blue: below baseline).

Figure 4. Head-wise analyses in different attention zones for SQuAD and CMRC 2018. 

Figure 5. Analyses of different question types for SQuAD. The number of each question type (in order): what (6073), how (1389), who (1377), when (864), which (747), where (508), why (158). 

Figure 6. Comparison of the development and challenge set for CMRC 2018.

Figure 7. Analyses of base-level and large-level BERT, ELECTRA, ALBERT for SQuAD.

Figure 8. Attention maps for different attention heads in layer 11 and 12. Recall that the input is created by the concatenation of ``[CLS] Q [SEP] P [SEP]''. The darker line means a strong connection between two words.

Figure 9. Visualization of specific attention heads in layer 11 (upper) and 12 (lower) of SQuAD model. The four boxes behind each word represents the values of each attention head (the higher value the darker).

\section*{Main Tables and Legends}

Table 1. Results on masking part of the attention map.

Table 2. Results on removing Top-$k$ and all attention scores. Only EM scores are reported.

Table 3. Pearson correlation of masking top-$k^\text{th}$ attention score. We report five-run average and its standard deviations.

\appendix
\section*{Star$\bigstar$Methods}\label{sec-star-methods}
\subsection*{Key Resources Table}
\begin{table}[h]
\small
\begin{center}
\begin{tabular}{l p{8cm} l}
\toprule
\bf Reagent or Resource & \bf Source & \bf Identifier \\
\midrule
SQuAD Training Set	& \url{https://rajpurkar.github.io/SQuAD-explorer/dataset/train-v1.1.json} & Version 1.1 \\
SQuAD Dev Set	& \url{https://rajpurkar.github.io/SQuAD-explorer/dataset/dev-v1.1.json} & Version 1.1 \\
CMRC 2018 Training Set & \url{https://github.com/ymcui/cmrc2018} & N/A \\
CMRC 2018 Dev Set & \url{https://github.com/ymcui/cmrc2018} & N/A \\
Python & \url{https://www.python.org} & Version 3.8 \\
TensorFlow & \url{https://tensorflow.org} & Version 1.15 \\
matplotlib & \url{https://matplotlib.org} & Version 3.4.0 \\
Captum & \url{https://github.com/pytorch/captum} & Version 0.4.0 \\
bertviz & \url{https://github.com/jessevig/bertviz} & Version 1.1.0 \\
MRC Model Analysis & \url{https://github.com/ymcui/mrc-model-analysis} & N/A \\
Chinese BERT-base & \url{https://storage.googleapis.com/bert_models/2018_11_03/chinese_L-12_H-768_A-12.zip} & N/A \\
English BERT-base-cased & \url{https://storage.googleapis.com/bert_models/2018_10_18/cased_L-12_H-768_A-12.zip} & N/A \\
English ALBERT-base & \url{https://storage.googleapis.com/albert_models/albert_base_v1.tar.gz} & N/A \\
English ALBERT-large & \url{https://storage.googleapis.com/albert_models/albert_large_v1.tar.gz} & N/A\\
English ELECTRA-base & \url{https://storage.googleapis.com/electra-data/electra_base.zip} & N/A \\
English ELECTRA-large & \url{https://storage.googleapis.com/electra-data/electra_large.zip} & N/A \\
\bottomrule
\end{tabular}
\end{center}
\end{table}

\subsection*{Resource Availability}
\subsubsection*{Lead Contact}
Further information and requests for resources and/or reagents should be directed to and will be fulfilled by the lead contact, Yiming Cui (ymcui@ir.hit.edu.cn).

\subsubsection*{Materials Availability}
This study did not generate new unique reagents.

\subsubsection*{Data and Code Availability}
\begin{itemize}
	\item Code: The source codes for the main experiments are publicly available on GitHub at \url{https://github.com/ymcui/mrc-model-analysis}.
	\item Dataset: All datasets used in this paper are publicly available, listed in the key resource table.
	\item Additional information: Any additional information required to reanalyze the data reported in this paper is available from the lead contact upon reasonable request.
\end{itemize}

\subsection*{Method Details}
\subsubsection*{Datasets}
In this paper, we mainly conduct our experiments on two span-extraction machine reading comprehension datasets.
\begin{itemize}
	\item {\bf SQuAD} \citep{rajpurkar-etal-2016}: This is the first span-extraction MRC dataset with over 100K samples. The dataset is constructed by English Wikipedia pages. SQuAD has been a leading benchmark in MRC research.
	\item {\bf CMRC 2018} \citep{cui-emnlp2019-cmrc2018}: This is also a span-extraction MRC dataset but in Chinese. The dataset is constructed by Chinese Wikipedia and with 10K human-annotated questions. Besides traditional train/dev/test splits, CMRC 2018 also contains a challenge set consisting of hard questions. 
\end{itemize}

\subsubsection*{Probing Method}

Pre-trained language model, such as BERT \citep{devlin-etal-2019-bert}, mainly comprises stacked multi-head self-attention layers with several dense layers. 
Given a hidden representation $\bm{H} \in \mathbb{R}^{n \times d}$ ($n$ for length of input and $d$ for hidden dimension), the model first uses three dense layers to transform $\bm{H}$ into query, key, and value representations.
\begin{gather}
	\bm{Q} = \bm{H} \bm{W}^\text{Q} ~,~ \bm{W}^\text{Q} \in \mathbb{R}^{d \times d} \\
	\bm{K} = \bm{H} \bm{W}^\text{K} ~,~ \bm{W}^\text{K} \in \mathbb{R}^{d \times d}  \\
	\bm{V} = \bm{H} \bm{W}^\text{V} ~,~ \bm{W}^\text{V} \in \mathbb{R}^{d \times d} 
\end{gather}
Then we calculate the dot product of query and key representations and apply softmax function to get the attention map $\bm{M} \in \mathbb{R}^{n \times n}$ ($d_a$ is the dimension of each attention head), indicating the correlations between each input token.
\begin{gather}
	\bm{M}' = \frac{1}{\sqrt{d_a}} \bm{Q} \bm{K}^\top \\
	\bm{M} = \mathrm{softmax}(\bm{M}')
\end{gather}
Finally, the dot product of attention matrix $\bm{M}$ and value representation $\bm{V}$ is calculated as the final self-attended representation $\bm{H}'$.
\begin{gather}
	\bm{H}' = \bm{M} \bm{V}
\end{gather}

To examine the effect of each attention zone, we perform masking on the attention matrix (before softmax activation) $\bm{M}'$. 
For example, if we choose to mask Q$^2$ (upper left part in Figure \ref{attention-zones}), the values in Q$^2$ zone will be set to a big negative value (in this paper, we set as -10000).
After the softmax function, these negative values will be normalized to values close to zero, demonstrating that this area is disabled.

\subsubsection*{Evaluation Metrics}

For span-extraction MRC tasks, there are two evaluation metrics: exact match (EM) and F1.
\begin{itemize}
	\item {\bf EM}: This is to measure the exact match between the prediction and the ground truth. An exact match will give a score of 1, otherwise 0.
	\item {\bf F1}: This is to measure the text overlap between the prediction and the ground truth. If there are more words overlapping, F1 will be close to 1, otherwise 0.
\end{itemize}

\subsubsection*{Hyperparameters and Detailed Setups}
The implementation is performed on the official fine-tuning script based on TensorFlow \citep{abadi2016tensorflow}.\footnote{https://github.com/google-research/bert}
All models are trained for three epochs with a universal initial learning rate of 3e-5 and batch size of 64.
We set other hyper-parameters as default. 
For visualizations, we use {\tt bertviz} \citep{bertviz} and {\tt captum} \citep{kokhlikyan2020captum}.
All experiments are carried out on Cloud TPUs v2 (64G HBM) or v3 (128G HBM), depending on the magnitude of the model.

\subsection*{Quantification and Statistical Analysis}

Each experiment was repeated five times with different random seeds (and led to different weight initializations) to make the analysis more robust.
We mainly report the average scores for five runs and report its standard deviation whenever necessary.

\bibliography{main}

\begin{thebibliography}{31}
\expandafter\ifx\csname natexlab\endcsname\relax\def\natexlab#1{#1}\fi
\providecommand{\url}[1]{\texttt{#1}}
\providecommand{\href}[2]{#2}
\providecommand{\path}[1]{#1}
\providecommand{\DOIprefix}{doi:}
\providecommand{\ArXivprefix}{arXiv:}
\providecommand{\URLprefix}{URL: }
\providecommand{\Pubmedprefix}{pmid:}
\providecommand{\doi}[1]{\href{http://dx.doi.org/#1}{\path{#1}}}
\providecommand{\Pubmed}[1]{\href{pmid:#1}{\path{#1}}}
\providecommand{\bibinfo}[2]{#2}
\ifx\xfnm\relax \def\xfnm[#1]{\unskip,\space#1}\fi
\bibitem[{Abadi et~al.(2016)Abadi, Barham, Chen, Chen, Davis, Dean, Devin,
  Ghemawat, Irving, Isard et~al.}]{abadi2016tensorflow}
\bibinfo{author}{Abadi, M.}, \bibinfo{author}{Barham, P.},
  \bibinfo{author}{Chen, J.}, \bibinfo{author}{Chen, Z.},
  \bibinfo{author}{Davis, A.}, \bibinfo{author}{Dean, J.},
  \bibinfo{author}{Devin, M.}, \bibinfo{author}{Ghemawat, S.},
  \bibinfo{author}{Irving, G.}, \bibinfo{author}{Isard, M.}, et~al.,
  \bibinfo{year}{2016}.
\newblock \bibinfo{title}{Tensorflow: A system for large-scale machine
  learning}, in: \bibinfo{booktitle}{12th {USENIX} Symposium on Operating
  Systems Design and Implementation ({OSDI} 16)}, pp.
  \bibinfo{pages}{265--283}.
\newblock \URLprefix
  \url{https://www.usenix.org/system/files/conference/osdi16/osdi16-abadi.pdf}.
\bibitem[{Bahdanau et~al.(2014)Bahdanau, Cho and Bengio}]{bahdanau-etal-2014}
\bibinfo{author}{Bahdanau, D.}, \bibinfo{author}{Cho, K.},
  \bibinfo{author}{Bengio, Y.}, \bibinfo{year}{2014}.
\newblock \bibinfo{title}{Neural machine translation by jointly learning to
  align and translate}.
\newblock \bibinfo{journal}{arXiv preprint arXiv:1409.0473} .
\bibitem[{{Barredo Arrieta} et~al.(2020){Barredo Arrieta},
  {D{\'\i}az-Rodr{\'\i}guez}, {Del Ser}, {Bennetot}, {Tabik}, {Barbado},
  {Garc{\'\i}a}, {Gil-L{\'o}pez}, {Molina}, {Benjamins}, {Chatila} and
  {Herrera}}]{xai-survey}
\bibinfo{author}{{Barredo Arrieta}, A.},
  \bibinfo{author}{{D{\'\i}az-Rodr{\'\i}guez}, N.}, \bibinfo{author}{{Del Ser},
  J.}, \bibinfo{author}{{Bennetot}, A.}, \bibinfo{author}{{Tabik}, S.},
  \bibinfo{author}{{Barbado}, A.}, \bibinfo{author}{{Garc{\'\i}a}, S.},
  \bibinfo{author}{{Gil-L{\'o}pez}, S.}, \bibinfo{author}{{Molina}, D.},
  \bibinfo{author}{{Benjamins}, R.}, \bibinfo{author}{{Chatila}, R.},
  \bibinfo{author}{{Herrera}, F.}, \bibinfo{year}{2020}.
\newblock \bibinfo{title}{Explainable artificial intelligence (xai): Concepts,
  taxonomies, opportunities and challenges toward responsible ai}.
\newblock \bibinfo{journal}{Information Fusion} \bibinfo{volume}{58},
  \bibinfo{pages}{82--115}.
\newblock \URLprefix
  \url{https://www.sciencedirect.com/science/article/pii/S1566253519308103},
  \DOIprefix\doi{https://doi.org/10.1016/j.inffus.2019.12.012}.
\bibitem[{Bastings and Filippova(2020)}]{bastings-filippova-2020-elephant}
\bibinfo{author}{Bastings, J.}, \bibinfo{author}{Filippova, K.},
  \bibinfo{year}{2020}.
\newblock \bibinfo{title}{The elephant in the interpretability room: Why use
  attention as explanation when we have saliency methods?}, in:
  \bibinfo{booktitle}{Proceedings of the Third BlackboxNLP Workshop on
  Analyzing and Interpreting Neural Networks for NLP},
  \bibinfo{publisher}{Association for Computational Linguistics},
  \bibinfo{address}{Online}. pp. \bibinfo{pages}{149--155}.
\newblock \URLprefix
  \url{https://www.aclweb.org/anthology/2020.blackboxnlp-1.14},
  \DOIprefix\doi{10.18653/v1/2020.blackboxnlp-1.14}.
\bibitem[{Clark et~al.(2020)Clark, Luong, Le and Manning}]{clark2020electra}
\bibinfo{author}{Clark, K.}, \bibinfo{author}{Luong, M.T.},
  \bibinfo{author}{Le, Q.V.}, \bibinfo{author}{Manning, C.D.},
  \bibinfo{year}{2020}.
\newblock \bibinfo{title}{{ELECTRA}: Pre-training text encoders as
  discriminators rather than generators}, in: \bibinfo{booktitle}{ICLR}.
\newblock \URLprefix \url{https://openreview.net/pdf?id=r1xMH1BtvB}.
\bibitem[{Cui et~al.(2021a)Cui, Che, Liu, Qin and Yang}]{chinese-bert-wwm}
\bibinfo{author}{Cui, Y.}, \bibinfo{author}{Che, W.}, \bibinfo{author}{Liu,
  T.}, \bibinfo{author}{Qin, B.}, \bibinfo{author}{Yang, Z.},
  \bibinfo{year}{2021}a.
\newblock \bibinfo{title}{Pre-training with whole word masking for chinese
  bert}.
\newblock \bibinfo{journal}{IEEE/ACM Transactions on Audio, Speech, and
  Language Processing} \bibinfo{volume}{29}, \bibinfo{pages}{3504--3514}.
\newblock \DOIprefix\doi{10.1109/TASLP.2021.3124365}.
\bibitem[{Cui et~al.(2017)Cui, Chen, Wei, Wang, Liu and Hu}]{cui-acl2017-aoa}
\bibinfo{author}{Cui, Y.}, \bibinfo{author}{Chen, Z.}, \bibinfo{author}{Wei,
  S.}, \bibinfo{author}{Wang, S.}, \bibinfo{author}{Liu, T.},
  \bibinfo{author}{Hu, G.}, \bibinfo{year}{2017}.
\newblock \bibinfo{title}{Attention-over-attention neural networks for reading
  comprehension}, in: \bibinfo{booktitle}{Proceedings of the 55th Annual
  Meeting of the Association for Computational Linguistics (Volume 1: Long
  Papers)}, \bibinfo{publisher}{Association for Computational Linguistics},
  \bibinfo{address}{Vancouver, Canada}. pp. \bibinfo{pages}{593--602}.
\newblock \URLprefix \url{https://aclanthology.org/P17-1055},
  \DOIprefix\doi{10.18653/v1/P17-1055}.
\bibitem[{Cui et~al.(2021b)Cui, Liu, Che, Chen and Wang}]{cui-etal-2021-expmrc}
\bibinfo{author}{Cui, Y.}, \bibinfo{author}{Liu, T.}, \bibinfo{author}{Che,
  W.}, \bibinfo{author}{Chen, Z.}, \bibinfo{author}{Wang, S.},
  \bibinfo{year}{2021}b.
\newblock \bibinfo{title}{{ExpMRC}: Explainability evaluation for machine
  reading comprehension}.
\newblock \bibinfo{journal}{arXiv preprint arXiv:2105.04126}
  \DOIprefix\doi{https://doi.org/10.48550/arXiv.2105.04126}.
\bibitem[{Cui et~al.(2022)Cui, Liu, Che, Chen and
  Wang}]{cui-etal-2022-teaching}
\bibinfo{author}{Cui, Y.}, \bibinfo{author}{Liu, T.}, \bibinfo{author}{Che,
  W.}, \bibinfo{author}{Chen, Z.}, \bibinfo{author}{Wang, S.},
  \bibinfo{year}{2022}.
\newblock \bibinfo{title}{Teaching machines to read, answer and explain}.
\newblock \bibinfo{journal}{IEEE/ACM Transactions on Audio, Speech, and
  Language Processing} \DOIprefix\doi{10.1109/TASLP.2022.3156789}.
\bibitem[{Cui et~al.(2019)Cui, Liu, Che, Xiao, Chen, Ma, Wang and
  Hu}]{cui-emnlp2019-cmrc2018}
\bibinfo{author}{Cui, Y.}, \bibinfo{author}{Liu, T.}, \bibinfo{author}{Che,
  W.}, \bibinfo{author}{Xiao, L.}, \bibinfo{author}{Chen, Z.},
  \bibinfo{author}{Ma, W.}, \bibinfo{author}{Wang, S.}, \bibinfo{author}{Hu,
  G.}, \bibinfo{year}{2019}.
\newblock \bibinfo{title}{A span-extraction dataset for {C}hinese machine
  reading comprehension}, in: \bibinfo{booktitle}{Proceedings of the 2019
  Conference on Empirical Methods in Natural Language Processing and the 9th
  International Joint Conference on Natural Language Processing
  (EMNLP-IJCNLP)}, \bibinfo{publisher}{Association for Computational
  Linguistics}, \bibinfo{address}{Hong Kong, China}. pp.
  \bibinfo{pages}{5886--5891}.
\newblock \DOIprefix\doi{10.18653/v1/D19-1600}.
\bibitem[{Devlin et~al.(2019)Devlin, Chang, Lee and
  Toutanova}]{devlin-etal-2019-bert}
\bibinfo{author}{Devlin, J.}, \bibinfo{author}{Chang, M.W.},
  \bibinfo{author}{Lee, K.}, \bibinfo{author}{Toutanova, K.},
  \bibinfo{year}{2019}.
\newblock \bibinfo{title}{{BERT}: Pre-training of deep bidirectional
  transformers for language understanding}, in: \bibinfo{booktitle}{Proceedings
  of the 2019 Conference of the North {A}merican Chapter of the Association for
  Computational Linguistics: Human Language Technologies, Volume 1 (Long and
  Short Papers)}, \bibinfo{publisher}{Association for Computational
  Linguistics}, \bibinfo{address}{Minneapolis, Minnesota}. pp.
  \bibinfo{pages}{4171--4186}.
\newblock \DOIprefix\doi{10.18653/v1/N19-1423}.
\bibitem[{Dhingra et~al.(2017)Dhingra, Liu, Yang, Cohen and
  Salakhutdinov}]{dhingra-etal-2017}
\bibinfo{author}{Dhingra, B.}, \bibinfo{author}{Liu, H.},
  \bibinfo{author}{Yang, Z.}, \bibinfo{author}{Cohen, W.},
  \bibinfo{author}{Salakhutdinov, R.}, \bibinfo{year}{2017}.
\newblock \bibinfo{title}{Gated-attention readers for text comprehension}, in:
  \bibinfo{booktitle}{Proceedings of the 55th Annual Meeting of the Association
  for Computational Linguistics (Volume 1: Long Papers)},
  \bibinfo{publisher}{Association for Computational Linguistics},
  \bibinfo{address}{Vancouver, Canada}. pp. \bibinfo{pages}{1832--1846}.
\newblock \URLprefix \url{https://aclanthology.org/P17-1168},
  \DOIprefix\doi{10.18653/v1/P17-1168}.
\bibitem[{Gunning(2017)}]{gunning2017explainable}
\bibinfo{author}{Gunning, D.}, \bibinfo{year}{2017}.
\newblock \bibinfo{title}{Explainable artificial intelligence ({XAI})}.
\newblock \bibinfo{journal}{Defense Advanced Research Projects Agency (DARPA),
  nd Web} \bibinfo{volume}{2}.
\bibitem[{Jain and Wallace(2019)}]{jain-wallace-2019-attention}
\bibinfo{author}{Jain, S.}, \bibinfo{author}{Wallace, B.C.},
  \bibinfo{year}{2019}.
\newblock \bibinfo{title}{{A}ttention is not {E}xplanation}, in:
  \bibinfo{booktitle}{Proceedings of the 2019 Conference of the North
  {A}merican Chapter of the Association for Computational Linguistics: Human
  Language Technologies, Volume 1 (Long and Short Papers)},
  \bibinfo{publisher}{Association for Computational Linguistics},
  \bibinfo{address}{Minneapolis, Minnesota}. pp. \bibinfo{pages}{3543--3556}.
\newblock \URLprefix \url{https://www.aclweb.org/anthology/N19-1357},
  \DOIprefix\doi{10.18653/v1/N19-1357}.
\bibitem[{Kadlec et~al.(2016)Kadlec, Schmid, Bajgar and
  Kleindienst}]{kadlec-etal-2016}
\bibinfo{author}{Kadlec, R.}, \bibinfo{author}{Schmid, M.},
  \bibinfo{author}{Bajgar, O.}, \bibinfo{author}{Kleindienst, J.},
  \bibinfo{year}{2016}.
\newblock \bibinfo{title}{Text understanding with the attention sum reader
  network}, in: \bibinfo{booktitle}{Proceedings of the 54th Annual Meeting of
  the Association for Computational Linguistics (Volume 1: Long Papers)},
  \bibinfo{publisher}{Association for Computational Linguistics},
  \bibinfo{address}{Berlin, Germany}. pp. \bibinfo{pages}{908--918}.
\newblock \URLprefix \url{https://aclanthology.org/P16-1086},
  \DOIprefix\doi{10.18653/v1/P16-1086}.
\bibitem[{Kokhlikyan et~al.(2020)Kokhlikyan, Miglani, Martin, Wang, Alsallakh,
  Reynolds, Melnikov, Kliushkina, Araya, Yan et~al.}]{kokhlikyan2020captum}
\bibinfo{author}{Kokhlikyan, N.}, \bibinfo{author}{Miglani, V.},
  \bibinfo{author}{Martin, M.}, \bibinfo{author}{Wang, E.},
  \bibinfo{author}{Alsallakh, B.}, \bibinfo{author}{Reynolds, J.},
  \bibinfo{author}{Melnikov, A.}, \bibinfo{author}{Kliushkina, N.},
  \bibinfo{author}{Araya, C.}, \bibinfo{author}{Yan, S.}, et~al.,
  \bibinfo{year}{2020}.
\newblock \bibinfo{title}{Captum: A unified and generic model interpretability
  library for pytorch}.
\newblock \bibinfo{journal}{arXiv preprint arXiv:2009.07896}
  \DOIprefix\doi{https://doi.org/10.48550/arXiv.2009.07896}.
\bibitem[{Kovaleva et~al.(2019)Kovaleva, Romanov, Rogers and
  Rumshisky}]{kovaleva-etal-2019-revealing}
\bibinfo{author}{Kovaleva, O.}, \bibinfo{author}{Romanov, A.},
  \bibinfo{author}{Rogers, A.}, \bibinfo{author}{Rumshisky, A.},
  \bibinfo{year}{2019}.
\newblock \bibinfo{title}{Revealing the dark secrets of {BERT}}, in:
  \bibinfo{booktitle}{Proceedings of the 2019 Conference on Empirical Methods
  in Natural Language Processing and the 9th International Joint Conference on
  Natural Language Processing (EMNLP-IJCNLP)}, \bibinfo{publisher}{Association
  for Computational Linguistics}, \bibinfo{address}{Hong Kong, China}. pp.
  \bibinfo{pages}{4365--4374}.
\newblock \URLprefix \url{https://www.aclweb.org/anthology/D19-1445},
  \DOIprefix\doi{10.18653/v1/D19-1445}.
\bibitem[{Lan et~al.(2020)Lan, Chen, Goodman, Gimpel, Sharma and
  Soricut}]{Lan2019ALBERT}
\bibinfo{author}{Lan, Z.}, \bibinfo{author}{Chen, M.},
  \bibinfo{author}{Goodman, S.}, \bibinfo{author}{Gimpel, K.},
  \bibinfo{author}{Sharma, P.}, \bibinfo{author}{Soricut, R.},
  \bibinfo{year}{2020}.
\newblock \bibinfo{title}{Albert: A lite bert for self-supervised learning of
  language representations}, in: \bibinfo{booktitle}{International Conference
  on Learning Representations (ICLR 2020)}.
\newblock \URLprefix \url{https://openreview.net/forum?id=H1eA7AEtvS}.
\bibitem[{Liu et~al.(2019)Liu, Ott, Goyal, Du, Joshi, Chen, Levy, Lewis,
  Zettlemoyer and Stoyanov}]{liu2019roberta}
\bibinfo{author}{Liu, Y.}, \bibinfo{author}{Ott, M.}, \bibinfo{author}{Goyal,
  N.}, \bibinfo{author}{Du, J.}, \bibinfo{author}{Joshi, M.},
  \bibinfo{author}{Chen, D.}, \bibinfo{author}{Levy, O.},
  \bibinfo{author}{Lewis, M.}, \bibinfo{author}{Zettlemoyer, L.},
  \bibinfo{author}{Stoyanov, V.}, \bibinfo{year}{2019}.
\newblock \bibinfo{title}{Roberta: A robustly optimized bert pretraining
  approach}.
\newblock \bibinfo{journal}{arXiv preprint arXiv:1907.11692}
  \DOIprefix\doi{https://doi.org/10.48550/arXiv.1907.11692}.
\bibitem[{Murdoch et~al.(2019)Murdoch, Singh, Kumbier, Abbasi-Asl and
  Yu}]{james-etal-2019}
\bibinfo{author}{Murdoch, W.J.}, \bibinfo{author}{Singh, C.},
  \bibinfo{author}{Kumbier, K.}, \bibinfo{author}{Abbasi-Asl, R.},
  \bibinfo{author}{Yu, B.}, \bibinfo{year}{2019}.
\newblock \bibinfo{title}{Definitions, methods, and applications in
  interpretable machine learning}.
\newblock \bibinfo{journal}{Proceedings of the National Academy of Sciences}
  \bibinfo{volume}{116}, \bibinfo{pages}{22071--22080}.
\newblock \URLprefix
  \url{https://www.pnas.org/doi/abs/10.1073/pnas.1900654116},
  \DOIprefix\doi{10.1073/pnas.1900654116}.
\bibitem[{Preechakul et~al.(2022)Preechakul, Sriswasdi, Kijsirikul and
  Chuangsuwanich}]{Preechakul2022}
\bibinfo{author}{Preechakul, K.}, \bibinfo{author}{Sriswasdi, S.},
  \bibinfo{author}{Kijsirikul, B.}, \bibinfo{author}{Chuangsuwanich, E.},
  \bibinfo{year}{2022}.
\newblock \bibinfo{title}{Improved image classification explainability with
  high-accuracy heatmaps}.
\newblock \bibinfo{journal}{iScience} \bibinfo{volume}{25}.
\newblock \URLprefix \url{https://doi.org/10.1016/j.isci.2022.103933},
  \DOIprefix\doi{10.1016/j.isci.2022.103933}.
\bibitem[{Rajpurkar et~al.(2018)Rajpurkar, Jia and
  Liang}]{rajpurkar-etal-2018-know}
\bibinfo{author}{Rajpurkar, P.}, \bibinfo{author}{Jia, R.},
  \bibinfo{author}{Liang, P.}, \bibinfo{year}{2018}.
\newblock \bibinfo{title}{Know what you don{'}t know: Unanswerable questions
  for {SQ}u{AD}}, in: \bibinfo{booktitle}{Proceedings of the 56th Annual
  Meeting of the Association for Computational Linguistics (Volume 2: Short
  Papers)}, \bibinfo{publisher}{Association for Computational Linguistics},
  \bibinfo{address}{Melbourne, Australia}. pp. \bibinfo{pages}{784--789}.
\newblock \DOIprefix\doi{10.18653/v1/P18-2124}.
\bibitem[{Rajpurkar et~al.(2016)Rajpurkar, Zhang, Lopyrev and
  Liang}]{rajpurkar-etal-2016}
\bibinfo{author}{Rajpurkar, P.}, \bibinfo{author}{Zhang, J.},
  \bibinfo{author}{Lopyrev, K.}, \bibinfo{author}{Liang, P.},
  \bibinfo{year}{2016}.
\newblock \bibinfo{title}{{SQuAD}: 100,000+ questions for machine comprehension
  of text}, in: \bibinfo{booktitle}{Proceedings of the 2016 Conference on
  Empirical Methods in Natural Language Processing},
  \bibinfo{publisher}{Association for Computational Linguistics}. pp.
  \bibinfo{pages}{2383--2392}.
\newblock \DOIprefix\doi{10.18653/v1/D16-1264}.
\bibitem[{Rudin(2019)}]{rudin-etal-2019}
\bibinfo{author}{Rudin, C.}, \bibinfo{year}{2019}.
\newblock \bibinfo{title}{Stop explaining black box machine learning models for
  high stakes decisions and use interpretable models instead}.
\newblock \bibinfo{journal}{Nature Machine Intelligence} \bibinfo{volume}{1},
  \bibinfo{pages}{206--215}.
\newblock \DOIprefix\doi{10.1038/s42256-019-0048-x}.
\bibitem[{Vaswani et~al.(2017)Vaswani, Shazeer, Parmar, Uszkoreit, Jones,
  Gomez, Kaiser and Polosukhin}]{vaswani2017attention}
\bibinfo{author}{Vaswani, A.}, \bibinfo{author}{Shazeer, N.},
  \bibinfo{author}{Parmar, N.}, \bibinfo{author}{Uszkoreit, J.},
  \bibinfo{author}{Jones, L.}, \bibinfo{author}{Gomez, A.N.},
  \bibinfo{author}{Kaiser, {\L}.}, \bibinfo{author}{Polosukhin, I.},
  \bibinfo{year}{2017}.
\newblock \bibinfo{title}{Attention is all you need}, in:
  \bibinfo{booktitle}{Advances in neural information processing systems}, pp.
  \bibinfo{pages}{5998--6008}.
\newblock \URLprefix
  \url{https://proceedings.neurips.cc/paper/2017/file/3f5ee243547dee91fbd053c1c4a845aa-Paper.pdf}.
\bibitem[{Vig(2019)}]{bertviz}
\bibinfo{author}{Vig, J.}, \bibinfo{year}{2019}.
\newblock \bibinfo{title}{A multiscale visualization of attention in the
  transformer model}, in: \bibinfo{booktitle}{Proceedings of the 57th Annual
  Meeting of the Association for Computational Linguistics: System
  Demonstrations}, \bibinfo{publisher}{Association for Computational
  Linguistics}, \bibinfo{address}{Florence, Italy}. pp.
  \bibinfo{pages}{37--42}.
\newblock \URLprefix \url{https://www.aclweb.org/anthology/P19-3007},
  \DOIprefix\doi{10.18653/v1/P19-3007}.
\bibitem[{Vuli{\'c} et~al.(2020)Vuli{\'c}, Ponti, Litschko, Glava{\v{s}} and
  Korhonen}]{vulic-etal-2020-probing}
\bibinfo{author}{Vuli{\'c}, I.}, \bibinfo{author}{Ponti, E.M.},
  \bibinfo{author}{Litschko, R.}, \bibinfo{author}{Glava{\v{s}}, G.},
  \bibinfo{author}{Korhonen, A.}, \bibinfo{year}{2020}.
\newblock \bibinfo{title}{Probing pretrained language models for lexical
  semantics}, in: \bibinfo{booktitle}{Proceedings of the 2020 Conference on
  Empirical Methods in Natural Language Processing (EMNLP)},
  \bibinfo{publisher}{Association for Computational Linguistics},
  \bibinfo{address}{Online}. pp. \bibinfo{pages}{7222--7240}.
\newblock \DOIprefix\doi{10.18653/v1/2020.emnlp-main.586}.
\bibitem[{Wiegreffe and Pinter(2019)}]{wiegreffe-pinter-2019-attention}
\bibinfo{author}{Wiegreffe, S.}, \bibinfo{author}{Pinter, Y.},
  \bibinfo{year}{2019}.
\newblock \bibinfo{title}{Attention is not not explanation}, in:
  \bibinfo{booktitle}{Proceedings of the 2019 Conference on Empirical Methods
  in Natural Language Processing and the 9th International Joint Conference on
  Natural Language Processing (EMNLP-IJCNLP)}, \bibinfo{publisher}{Association
  for Computational Linguistics}, \bibinfo{address}{Hong Kong, China}. pp.
  \bibinfo{pages}{11--20}.
\newblock \URLprefix \url{https://www.aclweb.org/anthology/D19-1002},
  \DOIprefix\doi{10.18653/v1/D19-1002}.
\bibitem[{Wu et~al.(2021)Wu, Arendt and Volkova}]{wu-etal-2021-evaluating}
\bibinfo{author}{Wu, W.}, \bibinfo{author}{Arendt, D.},
  \bibinfo{author}{Volkova, S.}, \bibinfo{year}{2021}.
\newblock \bibinfo{title}{Evaluating neural model robustness for machine
  comprehension}, in: \bibinfo{booktitle}{Proceedings of the 16th Conference of
  the European Chapter of the Association for Computational Linguistics: Main
  Volume}, \bibinfo{publisher}{Association for Computational Linguistics},
  \bibinfo{address}{Online}. pp. \bibinfo{pages}{2470--2481}.
\newblock \URLprefix \url{https://aclanthology.org/2021.eacl-main.210},
  \DOIprefix\doi{10.18653/v1/2021.eacl-main.210}.
\bibitem[{Xiao(2018)}]{xiao2018bertservice}
\bibinfo{author}{Xiao, H.}, \bibinfo{year}{2018}.
\newblock \bibinfo{title}{bert-as-service}.
\newblock
  \bibinfo{howpublished}{\url{https://github.com/hanxiao/bert-as-service}}.
\bibitem[{Yang et~al.(2018)Yang, Qi, Zhang, Bengio, Cohen, Salakhutdinov and
  Manning}]{yang-etal-2018-hotpotqa}
\bibinfo{author}{Yang, Z.}, \bibinfo{author}{Qi, P.}, \bibinfo{author}{Zhang,
  S.}, \bibinfo{author}{Bengio, Y.}, \bibinfo{author}{Cohen, W.},
  \bibinfo{author}{Salakhutdinov, R.}, \bibinfo{author}{Manning, C.D.},
  \bibinfo{year}{2018}.
\newblock \bibinfo{title}{{H}otpot{QA}: A dataset for diverse, explainable
  multi-hop question answering}, in: \bibinfo{booktitle}{Proceedings of the
  2018 Conference on Empirical Methods in Natural Language Processing},
  \bibinfo{publisher}{Association for Computational Linguistics},
  \bibinfo{address}{Brussels, Belgium}. pp. \bibinfo{pages}{2369--2380}.
\newblock \URLprefix \url{https://www.aclweb.org/anthology/D18-1259},
  \DOIprefix\doi{10.18653/v1/D18-1259}.

\end{thebibliography}

\end{document}